%% file: powerkernel_icmla2015.tex
\documentclass[conference]{IEEEtran}
\usepackage{amsthm}

\usepackage{url}
\usepackage{amsmath}
\usepackage{xspace}
\usepackage{graphicx}
\usepackage{numprint}
\usepackage{paralist}
\usepackage{booktabs}
\usepackage{placeins}
\usepackage{balance}

\newcommand{\sgn}{\operatorname{sgn}}

\usepackage{amssymb,amsbsy}
\usepackage{amsfonts}

\title{Complex Decomposition of the Negative Distance Kernel}
\author{\IEEEauthorblockN{Tim vor der Br{\"u}ck}
\IEEEauthorblockA{CC Distributed Secure Software Systems\\Luzerne University of\\Applied Sciences and Arts\\
tim.vorderbrueck@hslu.ch
}
\and
\IEEEauthorblockN{Steffen Eger and Alexander Mehler}
\IEEEauthorblockA{Text Technology Lab\\Goethe University Frankfurt am Main\\
\{steeger,amehler\}@em.uni-frankfurt.de}
}
\balance 
\begin{document}

© 2015 IEEE. Personal use of this material is permitted. Permission from IEEE must be obtained for all other uses, in any current or future media, including reprinting/republishing this material for advertising or promotional purposes, creating new collective works, for resale or redistribution to servers or lists, or reuse of any copyrighted component of this work in other works.

\maketitle
\begin{abstract}
A \textit{Support Vector Machine} (SVM) has become a very popular machine learning method for text classification. One reason for this relates to the range of existing kernels which allow for classifying data that is not linearly separable.
The linear, polynomial and RBF (Gaussian \textit{Radial Basis Function}) kernel are commonly used and serve as a basis of comparison in our study.
We show how to derive the primal form
of the quadratic \textit{Power Kernel} (PK) -- also called the \textit{Negative Euclidean Distance Kernel} (NDK) -- by means of complex numbers.
We exemplify the NDK in the framework of text categorization using the \textit{Dewey Document Classification} (DDC) as the target scheme. 
Our evaluation shows that the power kernel produces F-scores that are comparable to the reference kernels, but is -- except for the linear kernel -- faster to compute.
Finally, we show how to extend the NDK-approach by including the Mahalanobis distance.
\end{abstract}

%
\section{Introduction}

An SVM has become a very popular machine learning method for text classification \cite{Joachims:2002}. 
One reason for its popularity relates to the availability of a wide range of kernels including the linear, polynomial and RBF (Gaussian radial basis function) kernel. 
This paper derives a decomposition of the quadratic \textit{Power Kernel} (PK) using complex numbers and applies it in the area of text classification. 
Our evaluation shows that the NDK produces F-scores which are comparable to those produced by the latter reference kernels while being faster to compute -- except for the linear kernel.
This evaluation refers to the \textit{Dewey Document Classification} DDC \cite{oclc12} as the target scheme and compares our NDK-based classifier with two competitors described in \cite{loesch11}, \cite{mehler_waltinger09} and \cite{waltinger_etal12}, respecstively.

An SVM is a method for supervised classification introduced by \cite{vapnik98}. 
It determines a hyperplane that allows for the best
possible separation of the input data. 
(Training) data on the same side of the hyperplane is required to be mapped to the same class label. 
Furthermore, the margin of the hyperplane that is defined by the vectors located closest to the hyperplane is maximized. 
The vectors on the margin are called \textit{Support Vectors} (SV).
The decision function $dc: \mathbb{R}^n\rightarrow \{1,0,-1\}$, which maps a vector to its predicted class, is given by: 
$\mathop{dc}(\mathbf{x}):=\sgn(\langle \mathbf{w},\mathbf{x} \rangle +b)$
where $\sgn: \mathbb{R}\rightarrow \{1,0,-1\}$ is the Signum function; the vector $\mathbf{w}$ and the constant $b$ are determined by means of SV optimization. 
If $\mathop{dc}(\mathbf{x})=0$ then the vector is located directly on the hyperplane and no decision regarding both classes is possible.

The decision function is given in the primal form. It can be converted into the corresponding \textit{dual form}: 
$\mathop{dc}(\mathbf{x})=\sgn(\sum_{j=1}^m y_j \alpha_j \langle \mathbf{x},\mathbf{x}_j \rangle +b)$ where
$\alpha_j$ and $b$ are constants determined by the SV optimization and $m$ is the number of SVs $\mathbf{x}_j$ (the input vector has to be compared only with these SVs).
The vectors that are located on the wrong side of the hyperplane, that is, the vectors which prevent a perfect fit of SV optimization, are also considered as SVs. 
Thus, the number of SVs can be quite large.
The scalar product, which is used to estimate the  similarity of feature vectors, can be generalized to a kernel function. 
A kernel function $K$ is a similarity function of two vectors such that the matrix of kernel values is symmetric and positive semidefinite. 
The kernel function only appears in the dual form. 
The decision function is given as follows:
$\mathit{dc}(\mathbf{x}):=\sgn (\sum_{j=1}^m y_j \alpha_j K(\mathbf{x},\mathbf{x}_j) +b)$.
Let $\Phi: \mathbb{R}^n\rightarrow \mathbb{R}^l$ be a function that transforms a vector into another vector  mostly of higher dimensionality with $l\in \mathbb{N}\cup\{\infty\}$.
It is chosen in such a way that the kernel function can be represented by: 
$K(\mathbf{x}_1,\mathbf{x}_2)=\langle \Phi(\mathbf{x}_1), \Phi(\mathbf{x}_2) \rangle$.
Thus, the decision function in the primal form is given by:
\begin{equation}
\mathop{dc}(\mathbf{x}):=\sgn(\langle \mathbf{w},\Phi(\mathbf{x}) \rangle +b)
\label{eq:decision_primal}
\end{equation}

\noindent Note that $\Phi$ is not necessarily uniquely defined. 
Furthermore, $\Phi$ might convert the data into a very high dimensional space. 
In such cases, the dual form should be used for the optimization process as well as for the 
classification of previously unseen data. 
One may think that the primal form is not needed. However, it has one advantage:
if the normal vector of the hyperplane is known, a previously unseen vector can be classified just by applying the Signum function, $\Phi$ and a scalar multiplication. This is often much faster than computing the kernel function for previously unseen vectors and each SV as required when using the dual form.


The most popular kernel is the scalar product, also called the linear kernel. 
In this case, the transformation function $\Phi$ is the identity function:
$K_{\mathit{lin}}(\mathbf{x}_1,\mathbf{x}_2):=\langle \mathbf{x}_1,\mathbf{x}_2 \rangle$.
Another popular kernel function is the RBF (Gaussian Radial Basis Function), given by:
$K_r(\mathbf{x}_1,\mathbf{x}_2):=e^{-\gamma ||\mathbf{x}_1-\mathbf{x}_2||^2}$
where $\gamma\in \mathbb{R}$, $\gamma>0$ is a constant that has to be manually specified. This kernel function can be represented by a function $\Phi_r$ that transforms the vector into infinite dimensional space \cite{hung12}.  
For reasons of simplicity, assume that $\mathbf{x}$ is a vector with only one component.
Then the transformation function $\Phi_r$ is given by:
\begin{equation}
\begin{split}
\Phi_r(x):=&e^{-\gamma x^2}[1,\sqrt{\frac{2\gamma}{1!}} x, \sqrt{\frac{(2\gamma)^2}{2!}} x^2,\sqrt{\frac{(2\gamma)^3}{3!}} x^3,\ldots]^\top
\end{split}
\end{equation}

Another often used kernel is the polynomial kernel:
\begin{equation}
K_p(\mathbf{x}_1,\mathbf{x}_2):=(a\langle \mathbf{x}_1,\mathbf{x}_2\rangle+c)^d
\end{equation}
For $d:=2$, $a=1$, $c=1$, and two vector components, the function $\Phi_p$
is given by \cite{manning_etal08}: $\Phi_p: \mathbb{R}^2\rightarrow \mathbb{R}^{6}$ with 
\begin{equation}
\Phi_p(\mathbf{x})=(1,x_1^2,\sqrt{2}x_1x_2,x_2^2,\sqrt{2}x_1,\sqrt{2}x_2) 
\end{equation}

In general, a vector of dimension $n$ is transformed by $\Phi_p$ into a vector of dimension $\binom{n+d}{d}$.
Thus, the polynomial kernel leads to a large number of dimensions in the target space. However, the number of dimensions is not infinite as in the case of the RBF kernel.

\section{The power kernel}

The PK is a conditionally positive definite kernel given by
\begin{equation}
K_s(\mathbf{x}_1,\mathbf{x}_2):=-||\mathbf{x}_1-\mathbf{x}_2||^p
\end{equation}
 for some $p\in \mathbb{R}$ \cite{souza10,boolchandani_sahula11,sahbi_fleuret02}. A kernel is called conditionally positive-definite if it is symmetric and satisfies the conditions \cite{schoelkopf_smola02}
\begin{equation}
\sum_{j,k=1}^n c_i\overline{c_j}K(\mathbf{x}_j,\mathbf{x}_k)\geq 0\ \forall c_{i}\in \mathbb{K} \mathrm{\ with\ }\sum_{i=1}^m c_i=0
\end{equation}
where $\overline{c_j}$ is the complex-conjugate  of  $c_j$.
We consider here a generalized form of the PK for $p:=2$ (also called NDK):
\begin{equation}
K_{pow}(\mathbf{x}_1,\mathbf{x}_2):=-a||\mathbf{x}_1-\mathbf{x}_2||^2+c
\label{eq:power_kernel}
\end{equation}
with $a,c\in \mathbb{R}$ and $a>0$.
The expression $-a||\mathbf{x}_1-\mathbf{x}_2||^2+c$ can also be written as: 
\begin{equation}
\begin{split}
&-a\langle (\mathbf{x}_1-\mathbf{x}_2), (\mathbf{x}_1-\mathbf{x}_2)\rangle+c =\\
&-a(\langle \mathbf{x}_1,\mathbf{x}_1 \rangle-2\langle \mathbf{x}_1, \mathbf{x}_2 \rangle +\langle \mathbf{x}_2,\mathbf{x}_2 \rangle)+c
\end{split}
\end{equation}
For deciding which class a previously unseen vector belongs to we can use the 
decision function in the dual form:

\begin{equation}
\label{eq:modified_dual}
\begin{split}
dc(\mathbf{x}):&=\sgn(\sum_{j=1}^m y_j\alpha _j K_{\mathit{pow}}(\mathbf{x},\mathbf{x}_j) +b) \\
&=\sgn(\sum_{j=1}^m y_j\alpha_j (-a||\mathbf{x}-\mathbf{x}_j||^2+c)+b)
\end{split}
\end{equation}

The decision function shown in formula~\eqref{eq:modified_dual} has the drawback that the previously unseen vector
has to be compared with each SV, which can be quite time consuming.
This can be avoided, if we reformulate formula~\eqref{eq:modified_dual} to:
\begin{equation}
\label{eq:modified_dual2}
\begin{split}
dc(\mathbf{x})=&\sgn(\sum_{j=1}^m y_j\alpha_j (-a\langle \mathbf{x},\mathbf{x}\rangle +2a\langle \mathbf{x},\mathbf{x}_j\rangle-\\
&a\langle \mathbf{x}_j,\mathbf{x}_j \rangle+c)+b)\\
=&\sgn(-a\sum_{j=1}^m y_j\alpha_j \langle \mathbf{x},\mathbf{x}\rangle +2a \sum_{j=1}^m y_j\alpha_j \langle \mathbf{x},\mathbf{x}_j\rangle+\\
&\sum_{j=1}^my_j\alpha_j (-a\langle \mathbf{x}_j,\mathbf{x}_j \rangle+c)+b)\\
=&\sgn(-a\langle \mathbf{x}, \mathbf{x}\rangle \sum_{j=1}^m y_j \alpha_j +2 \langle \mathbf{x},a \sum_{j=1}^m y_j \alpha_j \mathbf{x}_j\rangle- \\
 &a\sum_{j=1}^m y_j\alpha_j\langle \mathbf{x}_j,\mathbf{x}_j \rangle+c\sum_{i=1}^m y_j\alpha_j+b)
\end{split}
\end{equation}
With $\mathbf{z}:=a\sum_{j=1}^m y_j \alpha_j \mathbf{x}_j$, $\mathbf{u}:= a\sum_{j=1}^m y_j \alpha_j  \langle \mathbf{x}_j,\mathbf{x}_j \rangle$ and 
$c'=c\sum_{i=1}^my_j\alpha_j$, formula~\eqref{eq:modified_dual2} can be rewritten as:
\begin{equation}
dc(\mathbf{x})=\sgn(-a\langle \mathbf{x},\mathbf{x}\rangle\sum_{i=1}^m y_j\alpha_j  +2\langle \mathbf{x},\mathbf{z}\rangle -\mathbf{u}+c'+b)
\end{equation}
The expressions $\mathbf{u}$, $\mathbf{z}$, $(\sum_{i=1}^m y_j\alpha_j)$, and $c'$ are identical for every vector $\mathbf{x}$ and can be precomputed.
Note that there exists no primal form for the NDK based on real number vectors which is stated by the following proposition. \\ 
\textbf{Proposition 1}
Let $a$,$c\in \mathbb{R}$ with $a>0$ and 
$n,l \in \mathbb{N}$. Then there is no function $\Phi_{\mathit{re}}:\mathbb{R}^n\rightarrow \mathbb{R}^l$ ($\mathit{re}$ indicates that $\Phi_{\mathit{re}}$
operates on real numbers) with   
$\forall \mathbf{x}_1,\mathbf{x}_2\in \mathbb{R}^n: 
\langle \Phi_{\mathit{re}}(\mathbf{x}_1), \Phi_{\mathit{re}}(\mathbf{x}_2) \rangle= -a||\mathbf{x}_1-\mathbf{x}_2||^2+c\  $. \\ 
\begin{IEEEproof}
If such a function existed, then, for all $\mathbf{x}\in \mathbb{R}^n$,
\[
||\Phi_{\mathit{re}}(\mathbf{x})||^2=\langle \Phi_{\mathit{re}}(\mathbf{x}),\Phi_{\mathit{re}}(\mathbf{x})\rangle=-a\cdot 0+c=c
\]
which requires that $c\geq 0$ since the square of a real number cannot be negative. Now, consider $\mathbf{x},\mathbf{y}\in \mathbb{R}^n$
with \\ $||\mathbf{x}-\mathbf{y}||^2>\frac{2c}{a}\geq 0$. On the one hand, we have, by the Cauchy-Schwarz inequality,
\[ |\langle \Phi_{\mathit{re}}(\mathbf{x}),\Phi_{\mathit{re}}(\mathbf{y}) \rangle|\leq ||\Phi_{\mathit{re}}(\mathbf{x})||\cdot ||\Phi_{\mathit{re}}(\mathbf{y})||=\sqrt{c}\cdot \sqrt{c}=c.
\]
On the other hand, it holds that
\[ |-a||\mathbf{x}-\mathbf{y}||^2+c|=|a||\mathbf{x}-\mathbf{y}||^2-c|>2c-c=c, \]
a contradiction.
\end{IEEEproof}

Although no primal form and therefore no function $\Phi_{\mathit{re}}$ exists for real number vectors, such a function can be given if complex number vectors are used instead. In this case, the function $\Phi_c$ is defined with a real domain and a complex co-domain:
$\Phi_c: \mathbb{R}^n \rightarrow \mathbb{C}^{4n+1}$ and
\begin{equation}
\label{eq:phi}
\begin{split}
\Phi_c(\mathbf{x}):=&(\sqrt{a}(x_{1}^2-1),\sqrt{a}i,\sqrt{2a}x_{1},\sqrt{a}ix_{1}^2,\ldots,\\
&\sqrt{a}(x_{n}^2-1),\sqrt{a}i,\sqrt{2a}x_{n},\sqrt{a}ix_{n}^2,\sqrt{c})^\top
\end{split}
\end{equation}

Note that no scalar product can be defined for complex numbers that fulfills the usual conditions of bilinearity and positive-definiteness simultaneously\footnote{This can be verified by a simple calculation: Consider some vector $\mathbf{x}\neq \mathbf{0} \mathrm{\ and\ }\mathbf{x}\in \mathbb{C}^n$. Since $\langle . \rangle$ is positive definite: 
$\langle \mathbf{x}, \mathbf{x} \rangle>0,$ by bilinearity: $\langle \sqrt{-i}\mathbf{x},\sqrt{-i}\mathbf{x} \rangle$ = 
$-i \langle \mathbf{x},\mathbf{x} \rangle\not >0$).}.
 Thus, the bilinearity condition is dropped for the official definition and only sesquilinearity is required.
The standard scalar product is defined as the sum of the products of the vector components with the associated complex conjugated vector components of the other vector. Let $\mathbf{x}_1$,$\mathbf{x}_2\in \mathbb{C}^n$, then the scalar product is given
by \cite{beutelspacher10}:
$\langle \mathbf{x}_1,\mathbf{x}_2 \rangle:=\sum_{k=1}^n x_{1k}\overline{x_{2k}}$.
In contrast, we use a modified scalar product (marked by a ``$^*$'') where, analogously to the real vector definition,
 the products of the vector components are summated:
$\mathit{\langle^*} \mathbf{x}_1,\mathbf{x}_2 \rangle:=\sum_{k=1}^n x_{1k}x_{2k}$.
This product (not a scalar product in strict mathematical sense) is a bilinear form but no longer positive definite.
For real number vectors this modified scalar product is
identical to the usual definition.
With this modified scalar product we get
\begin{equation}
\label{eq:power_kernel2}
\begin{split}
&\langle^* \Phi_c(\mathbf{x}_1),\Phi_c(\mathbf{x}_2) \rangle \\
 = &-ax_{11}^2-ax_{21}^2+2ax_{11}x_{21}-\cdots-ax_{1n}^2-\\
&ax_{2n}^2+2ax_{1n}x_{2n}+c =  -a||\mathbf{x}_1-\mathbf{x}_2||^2+c
\end{split}
\end{equation}

which is just the result of the NDK.
The optimization can be done with the dual form. Thus, no complex number optimization is necessary.
For the decision on the class to which a data vector should be assigned we switch to the primal form.
The vector $\mathbf{w}\in \mathbb{C}^{4n+1}$ is calculated by:\\ 
$\mathbf{w}:=\sum_{j=1}^m \alpha_j y_j \Phi_c(\mathbf{x}_j)$
 for all SVs $\mathbf{x}_j\in \mathbb{R}^n$.
The decision function is then given by:
$\mathop{dc}(\mathbf{x}):=\sgn(\langle^* \mathbf{w},\Phi_c(\mathbf{x}) \rangle+b)$.
Note that 
the modified scalar product
 $\langle^* \mathbf{w},\Phi_c(\mathbf{x}) \rangle$  must be a real number. This
is stated in the following proposition. \\ 
\noindent
$\textbf{Proposition 2}$
Let $\mathbf{w}=\sum_{j=1}^m \alpha_j y_j \Phi_c(\mathbf{x}_j)$ with $\mathbf{x}_j\in \mathbb{R}^n$, $\alpha_j \in \mathbb{R}$, $y_j\in \{-1,1\}$, $j=1,\ldots,m, \Phi_c$ as defined in formula~\eqref{eq:phi} and $\mathbf{x}\in\mathbb{R}^n$. Then
 $\langle^* \mathbf{w}, \Phi_c(\mathbf{x})\rangle $ is a real number.
\begin{IEEEproof}
$\langle^* \mathbf{w}, \Phi_c(\mathbf{x}) \rangle$ is given by:
\begin{equation}
\begin{split}
\langle^* \mathbf{w},\Phi_c(\mathbf{x}) \rangle &=\langle^* \sum_{j=1}^m \alpha_j y_j \Phi_c(\mathbf{x}_j),\Phi_c(\mathbf{x}) \rangle \\
& = \sum_{j=1}^m \langle^* \alpha_j y_j \Phi_c(\mathbf{x}_j),\Phi_c(\mathbf{x}) \rangle  \hspace*{0.2cm} (\langle^* . \rangle\mathrm{\ is\ bilinear}) \\
& = \sum_{j=1}^m \alpha_j y_j \langle^* \Phi_c(\mathbf{x}_j), \Phi_c(\mathbf{x}) \rangle \\
& = \sum_{j=1}^m \alpha_j y_j (-a||\mathbf{x}_j-\mathbf{x}||^2+c)  \hspace*{0.2cm} \mathrm{(see\ form.~\eqref{eq:power_kernel2})}
\end{split}
\end{equation}

which is clearly a real number. 
\end{IEEEproof}
The NDK is related to the polynomial kernel since it can also be represented by a polynomial.  However, it has the advantage over the polynomial kernel that it is faster to compute, since the  number of dimensions in the target space grows only linearly and not exponentially with the number of dimensions in the original space  
\cite{souza10,boolchandani_sahula11}. 
It remains to show that the decision functions following the primal and dual form are also equivalent for the modified form of the scalar product.
This is stated in the follow proposition: \\ 
\textbf{Proposition 3} Let $\mathbf{x}, \mathbf{x}_1,\ldots,\mathbf{x}_m\in \mathbb{R}^n$, $\mathrm{\alpha}\in \mathbb{R}^m$, $y\in \{-1,1\}^m$,  $\textbf{w}:=\sum_{j=1}^m \alpha_j y_j \Phi_c(\mathbf{x}_j)$ and 
 $\langle^*\Phi_c(\mathbf{z}_1),\Phi_c(\mathbf{z}_2)\rangle=K(\mathbf{z}_1,\mathbf{z_2})\  \forall \mathbf{z}_1,\mathbf{z}_2\in \mathbb{R}^n$.   
Then\\ $\sgn(\langle^* \mathbf{w},\Phi_c(\mathbf{x}) \rangle+b)=\sgn(\sum_{j=1}^m \alpha_j y_j K(\mathbf{x},\mathbf{x}_j)+b)$.  
\begin{IEEEproof}
\begin{equation}
\begin{split}
\hspace*{0.2cm}&\sgn(\langle^* \mathbf{w},\Phi_c(\mathbf{x}) \rangle+b)\\
= &\sgn(\langle^* \sum_{j=1}^m \alpha_j y_j \Phi_c(\mathbf{x}_j),\Phi_c(\mathbf{x}) \rangle+b) \\
= &\sgn(\sum_{j=1}^m \langle^* \alpha_j y_j \Phi_c(\mathbf{x}_j),\Phi_c(\mathbf{x}) \rangle+b) \hspace*{0.4cm}  \mathrm{(\langle^* .\rangle \ is\ bilinear) }\\
 =& \sgn(\sum_{j=1}^m\alpha_j y_j \langle^*  \Phi_c(\mathbf{x}_j), \Phi_c(\mathbf{x}) \rangle+b) \\
\end{split}
\end{equation}
\begin{equation}
\begin{split}
=& \sgn(\sum_{j=1}^m\alpha_j y_j K(\mathbf{x}_j,\mathbf{x})+b) \\
 =& \sgn(\sum_{j=1}^m\alpha_j y_j K(\mathbf{x},\mathbf{x}_j)+b)\hspace*{1.2cm} (K \mathrm{\ is\ symmetric)}
\end{split}
\end{equation}
\end{IEEEproof}


Normally, feature vectors for document classification represent the weighted occurrences of lemma or word forms \cite{Sebastiani02}.
Thus, such a vector contains a large number of zeros and is therefore usually considered sparse. In this case, the computational complexity of the
scalar product can be reduced from $\mathcal{O}(n)$ (where $n$ is the number of vector components) to some constant runtime, which is the average number of non-zero vector
components. Let $I_1$ be the set of indices of non-zero entries of vector $\mathbf{x}_1$ 
$(I_1 = \{ k \in \{1,\ldots,n\}: x_{1k} \neq 0 \})$
and $I_2$ be analogously defined for vector $\mathbf{x}_2$.
The scalar product of both vectors can then be computed by $\sum_{k\in(I_1\cap I_2)} x_{1k} \cdot x_{2k} $. 
Let us now consider the case that both vectors are transformed to complex numbers before the scalar multiplication. In this case, the 
modified scalar product 
\begin{equation}
\begin{split}
&  \langle^* \Phi_c(\mathbf{x}_{1}), \Phi_c(\mathbf{x}_{2}) \rangle = \langle (\phi(x_{11}),\ldots, \phi(x_{1n})) , \\
& (\phi(x_{21}), \ldots, \phi(x_{2n})) \rangle+c 
\end{split}
\end{equation}
 is considered where
$\phi(x_{k})$ denotes the transformation of a single real vector component to a complex number vector  and is defined as:
\begin{equation}
\begin{split}
\phi: & \mathbb{R} \rightarrow \mathbb{C}^4,\ \phi(x_{k}):=(\sqrt{a}(x_{k}^2-1),\sqrt{a}i,\sqrt{2a}x_{k},\sqrt{a}ix_{k}^2) 
\end{split}
\end{equation}

Note that the partial modified scalar product $\langle^* \phi(x_{1k}),   \phi(x_{2k}) \rangle$ can be 
non-zero, if at least one of the two vector components $x_{1k}$ and $x_{2k}$ is non-zero, which is easy to see:
\begin{equation}
\begin{split}
&\langle^* \phi(x_{k}),\phi(0)\rangle=\sqrt{a}(x_{k}^2-1)\cdot 
 \sqrt{a}(-1) + (-1)a  \\
+&\sqrt{2a}x_{k}\cdot 0 + \sqrt{a}ix_{k}^2\cdot 0  =a-ax^2_{k}-a=-ax^2_{k}
\end{split}
\end{equation}
Only if both vector components are zero, one can be sure that the result is also zero:
\begin{equation}
\langle^* \phi(0),\phi(0) \rangle= (-\sqrt{a})(-\sqrt{a})+ai\cdot i+0+0=a-a=0
\end{equation}

Thus, the sparse data handling of two transformed complex vectors has to be modified in such a way that vector components associated with 
the union and not the intersection of non-zero indices are considered for multiplication:
\begin{equation}
\langle^* \mathbf{x}_1,\mathbf{x}_2 \rangle=\sum_{k\in(I_1\cup I_2)}\langle^* \phi(x_{1k}), \phi(x_{2k})\rangle +c
\end{equation}


A further advantage of the NDK is that the Mahalanobis distance \cite{mahalanobis36} 
can easily be integrated, which is shown in the remainder of this section.
Each symmetric matrix $\mathbf{A}$ can be represented by the product $\mathbf{V}^{-1}\mathbf{D}\mathbf{V}$ where $\mathbf{D}$ is a diagonal matrix with the eigenvalues of $\mathbf{A}$ on its diagonals.
The square root of a matrix $\mathbf{A}$ is then defined as $\sqrt{\mathbf{A}}:=\mathbf{V}^{-1}\mathbf{D}^{0.5}\mathbf{V}$ where $\mathbf{D}^{0.5}$ is the matrix with the square root of the eigenvalues of $\mathbf{A}$ on its 
diagonal. It is obvious that $\sqrt{\mathbf{A}}\cdot \sqrt{\mathbf{A}}=\mathbf{A}$. \\
\textbf{Proposition 4}:
Let $\mathbf{x}$, $\mathbf{z}\in \mathbb{R}^n, a:=1, c:=0$,  then \\$\langle^* \Phi_c(\sqrt{\mathbf{Cov}^{-1}}\mathbf{x}),\Phi_c(\sqrt{\mathbf{Cov}^{-1}}\mathbf{z})\rangle =-\mathit{MH}(\mathbf{x},\mathbf{z})^2$ \\
where $\mathit{MH}(\mathbf{x},\mathbf{z})$ denotes the Mahalanobis distance $\sqrt{(\mathbf{x}-\mathbf{z})^T \mathbf{Cov}^{-1} (\mathbf{x}-\mathbf{z})}$ and  $\mathbf{Cov}$ denotes the covariance matrix between the feature values of the entire training data set.

\begin{IEEEproof}
We have that $(\mathbf{with\ }\mathbf{C}:=\sqrt{\mathbf{Cov}^{-1}})$:
\begin{equation}
\begin{split}
&\langle^*\Phi_c(\mathbf{C}\mathbf{x}),\Phi_c(\mathbf{C}\mathbf{z})\rangle \\
=&-||\mathbf{C}\mathbf{x}-\mathbf{C}\mathbf{z}||^2 \hspace*{1cm}(\mathrm{see\ formula}~\eqref{eq:power_kernel2}) \\ 
=&-(\mathbf{C}\mathbf{x}-\mathbf{C} \mathbf{z})^{\top}(\mathbf{C}\mathbf{x}-\mathbf{C}\mathbf{z}) \\
=&-(\mathbf{C}(\mathbf{x}-\mathbf{z}))^{\top}(\mathbf{C}(\mathbf{x}-\mathbf{z})) \\
=&-(\mathbf{x}-\mathbf{z})^{\top}\sqrt{\mathbf{Cov}^{-1}}^{\top}\sqrt{\mathbf{Cov}^{-1}}(\mathbf{x}-\mathbf{z}) \\
=&-(\mathbf{x}-\mathbf{z})^{\top}\mathbf{Cov}^{-1}(\mathbf{x}-\mathbf{z})
\end{split}
\end{equation}
(since the covariance matrix is symmetric and the\\
inverse and square root of a symmetric matrix is\\
also symmetric)\\
\end{IEEEproof}

This proposition shows that the NDK is just the negative square of the Mahalanobis distance for some vectors $\mathbf{\hat x}$ and $\mathbf{\hat z}$ if $\mathbf{\hat x}$ 
is set to $\sqrt{\mathbf{Cov}^{-1}}\mathbf{x}$ (analogously for $\mathbf{\hat z}$), 
$a$ is set to 1 and $c$ is set to 0.
Furthermore, this proposition shows that we can easily extend our primal form to integrate the Mahalanobis distance. For that,
we define a function $\tau: \mathbb{R}^n \rightarrow \mathbb{R}^n $ as follows: $\tau(\mathbf{x})= \sqrt{\mathbf{Cov}^{-1}}\mathbf{x}$. With
 $\Phi_m$:=$\Phi_c\circ \tau$ we have:
\begin{equation}
\langle^* \Phi_m(\mathbf{x}),\Phi_m(\mathbf{z}) \rangle= (\mathit{MH}(\mathbf{x},\mathbf{z}))^2 
\end{equation}
 which shows that we have indeed derived a primal form.


We use the NDK for text classification where  
a text is automatically labeled regarding its main topics
(i.e., categories), 
employing the DDC as one of the leading classification schemes in digital libraries \cite{oclc12}. 
%
Documents are classified regarding the 10 top-level  categories of the DDC.
To this end, we use training sets of documents for which the DDC categories have already been assigned.
Lexical features are extracted from this training set and made input to an SVM library (i.e., $libsvm$ \cite{chung_lin01}) in order to assign one or more DDC categories to previously unseen texts. 
Features of instance documents are computed by means of the geometric mean of the tfidf values and the GSS coefficients of their lexical constituents \cite{joachims00,salton_buckley98}.

\begin{table*}
\caption{Precision / recall of different kernels.}
\label{tab:precision_recall1}
\centering
\begin{tabular}{crr|rr|rr|rr|rr}
\toprule
&\multicolumn{2}{c}{NDK}&\multicolumn{2}{c}{RBF}&\multicolumn{2}{c}{Linear}&\multicolumn{2}{c}{Square}&\multicolumn{2}{c}{Cubic}\\
Cat.&Prec.&Rec.&Prec.&Rec.&Prec.&Rec.&Prec.&Rec.&Prec.&Recall\\
\midrule
0&0.845&0.777&0.867&0.731&0.747&0.853 &0.831&0.772 &  0.789&0.817\\
1&0.768&0.740&0.775&0.682&0.726&0.745 &0.722&0.771 & 0.712&0.734\\
2&0.909&0.833&0.892&0.857&0.905&0.847 &0.873&0.882 & 0.935& 0.635\\
3&0.764&0.497&0.679&0.571&0.545&0.640 &0.716&0.508 & 0.708& 0.487\\
4&0.852&0.529&0.870&0.388&0.863&0.490 &0.441&0.146 & 0.000& 0.029\\
5&0.700&0.783&0.794&0.626&0.825&0.675 &0.744&0.660 & 0.697& 0.793\\
6&0.682&0.685&0.718&0.625&0.699&0.685 &0.670&0.700 & 0.744& 0.595\\
7&0.680&0.741&0.675&0.652&0.602&0.751 &0.722&0.697 & 0.843& 0.532\\
8&0.657&0.720&0.586&0.785&0.626&0.774 &0.670&0.677 & 0.752& 0.457\\
9&0.701&0.752&0.680&0.670&0.668&0.723 &0.775&0.636& 0.831& 0.335 \\ 
\midrule
All&0.756&0.706&0.754&0.659&0.721&\bf{0.718}&0.716&  0.645&\bf{0.801} &  0.542\\
\bottomrule
\end{tabular}
\end{table*}

\section{Evaluation}

The evaluation and training was done using \numprint{4000} German documents, containing in total \numprint{114887606} words and \numprint{9643022} sentences requiring a storage space of 823.51 MB.
There are 400 documents of each top-level DDC category in the corpus. \numprint{2000} of the texts were used for training, \numprint{2000} for evaluation. The corpus consists of texts annotated according to the OAI (\textit{Open Archive Initiative}) and is presented in \cite{Loesch:Waltinger:Horstmann:Mehler:2011}. 
We tested the correctness of category assignment for the 10 top-level DDC categories. 
Each document is mapped at least to one DDC category. 
Multiple assignments are possible.   
Precision, recall and F-scores were computed for each DDC category using the NDK, the square (polynomial kernel of degree 2), the cubic (polynomial kernel of degree 3), the RBF kernel and the linear kernel 
(see~Tables~\ref{tab:precision_recall1} and \ref{tab:f_measure}). The free parameters of the square, the cubic and the RBF kernel are determined by means of a grid search on an independent data set. On the same held-out dataset, we adjusted the SVM-threshold parameter $b$ to optimize the F-scores.
The time (on an Intel Core i7) required to obtain the classifications was determined (see Table~\ref{tab:runtime1} for the real / complex NDK, the RBF, square, and linear kernel). The time needed for building the model was not measured because of being required only once and therefore being irrelevant for online text classification. 

The F-score for the NDK is higher than the F-scores of all other kernels and faster to compute except for the linear kernel -- obviously, the classification using the primal form of the linear kernel is faster than the one using the NDK. 
Furthermore, the complex decomposition of the NDK leads to a considerable acceleration of the classification process compared to its dual form and should therefore be preferred.  

We conducted a second experiment where we compared the linear kernel with the NDK now using text snippets (i.e., abstracts instead of full texts) taken once more from our OAI corpus. 
This application scenario is more realistic for digital libraries that mainly harvest documents by means of their meta data and, therefore, should also be addressed by text classifiers.
The kernels were trained on a set of \numprint{19000} abstracts\footnote{Note that in this evaluation we employed the tfidf score for weighting only, since the use of the GSS coefficient didn't lead to any improvement in F-score.}. This time, the categories are not treated independently of each other. Instead, the classifier selected all categories with signed distance values from the SVM hyperplane that are greater or equal to zero. If all distance values are negative, the category with the largest (smallest absolute) distance was chosen. In this experiment, the linear kernel performed better than the NDK. Using three samples of 500 texts, the F-scores of the linear kernel are (macro-averaged over all main 10 DDC categories):  
0.753, 0.735, 0.743,
and of the NDK: 
0.730, 0.731, 0.737. 
This result indicates that the heuristic of preferring the categories of highest signed distance to the hyperplane is not optimal for the NDK.
We compared our classifier with two other systems, the DDC classifier of \cite{waltinger_etal12} and of \cite{loesch11}. 
\cite{waltinger_etal12} reaches an F-score of 0.502, 0.457 and 0.450 on these three samples.
The F-scores of the DDC-classifier of \cite{loesch11} are 0.615, 0.604, and 0.574.
Obviously, our classifier outperforms these competitors. 

\begin{figure}
\centering
\input{pics/histogram}
\caption{Histogram of the distribution of categories for the Reuters corpus.}
\label{fig:histogram}
\end{figure}
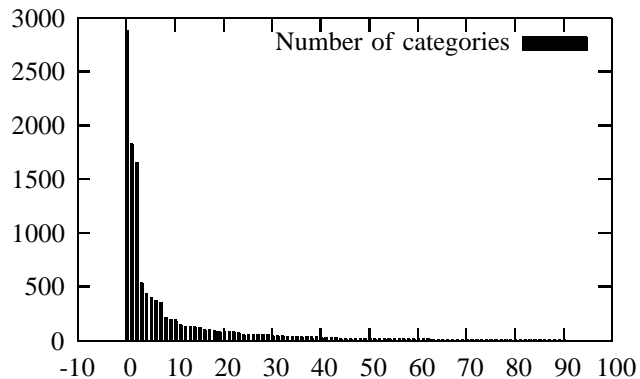

Finally, we evaluated the NDK on the Reuters 21578 corpus\footnote{URL: \url{http://www.daviddlewis.com/resources/testcollections/reuters21578}} (Lewis split). This test set is very challenging since 35 of all 93 categories occurring in this split have 10 or less training examples. Furthermore, several texts are intentionally assigned to none of the Reuters categories. We created an artificial category for all texts that are not assigned in this sense. 
The histogram of category assignments is displayed in Figure~\ref{fig:histogram}.
The F-scores for the Reuters corpus are given in Table~\ref{tab:eval_reuters}. We modified the training data by filling up instances for every category by randomly selecting positive instances such that the ratio of positive examples to all instances are the same for all categories. Hence, in most category samples some  training examples were used more than once. 
This approach prevents from preferring categories due to their multitude of training examples. 
The F-score of the NDK is lower than the highest macro-averaging F-score obtained by the linear kernel, but outperforms the square and the cubic kernel. Again, the parameters of the NDK, the square, the cubic and the RBF kernels were determined by means of a grid search. 

\section{Conclusion and future work}

We derived a primal form of the NDK by means of complex numbers. In this way, we obtained a much simpler representation compared to the modified dual form. We showed that the primal form (and in principle also the modified dual form) can speed up text classification considerably. The reason is that it does not require to compare input vectors with all support vectors. Our evaluation showed that the F-scores of the NDK are competitive regarding all other kernels tested here while the NDK consumes less time than the polynomial and the RBF kernel. 
We have also shown that the NDK performs better than the linear kernel when using full texts rather than text snippets.
Whether this is due to problems of feature selection/expansion or a general characteristic of this kernel (in the sense of being negatively affected by ultra-sparse features spaces), will be examined in future work.
Additionally, we plan to examine the PK with exponents larger than two, to investigate under which prerequisites the PK performs well and to evaluate the extension of the NDK that includes the Mahalanobis distance.

\begin{table}
\centering
\caption{F-scores of different kernels evaluated by means of the OAI corpus of \cite{Loesch:Waltinger:Horstmann:Mehler:2011}.}
\label{tab:f_measure}
\begin{tabular}{crrrrr}
\toprule
Cat.&NDK&Square&Cubic&RBF&Linear\\
\midrule
0&0.810&0.800&   0.803 &0.793&0.796\\
1&0.753&0.746&   0.723 &0.726&0.735\\
2&0.869&0.877&   0.757 &0.874&0.875\\
3&0.603&0.594&   0.577 &0.621&0.589\\
4&0.653&0.219&   0.057 &0.537&0.625\\
5&0.740&0.700&   0.742 &0.700&0.743\\
6&0.683&0.685&   0.661 &0.668&0.692\\
7&0.710&0.709&   0.652 &0.663&0.668\\
8&0.687&0.674&   0.569 &0.671&0.692\\
9&0.726&0.699&   0.478 &0.675&0.695\\
\midrule
All&\bf{0.723}&0.670&0.602&0.693&0.711\\
\bottomrule
\end{tabular}
\end{table}

\begin{table}
\centering
\caption{Time (in milliseconds) required  for the computation of the kernels on the OAI corpus.}
\label{tab:runtime1}
\begin{tabular}{crrrrrr}
\toprule
C.&NDK&NDK&Squ.&Cubic&RBF&Lin.\\
&prim.&dual&dual&dual&dual&dual\\
\midrule
0&\textbf{\numprint{6426}}&\numprint{46200}&\numprint{50256}&\numprint{50979}&\numprint{47246}&\numprint{44322}\\
1&\textbf{\numprint{9339}}&\numprint{98289}&\numprint{115992}&\numprint{132844}&\numprint{122751}&\numprint{93550}\\
2&\textbf{\numprint{5822}}&\numprint{48359}&\numprint{57603}&\numprint{67200}&\numprint{65508}&\numprint{46786}\\
3&\textbf{\numprint{23774}}&\numprint{134636}&\numprint{165604}&\numprint{177211}&\numprint{159897}&\numprint{111516}\\
4&\textbf{\numprint{10103}}&\numprint{118322}&\numprint{77634}&\numprint{93613}&\numprint{111153}&\numprint{103548}\\
5&\textbf{\numprint{32501}}&\numprint{104865}&\numprint{72772}&\numprint{135820}&\numprint{138107}&\numprint{96637}\\
6&\textbf{\numprint{24375}}&\numprint{105614}&\numprint{116892}&\numprint{120919}&\numprint{126498}&\numprint{95423}\\
7&\textbf{\numprint{19206}}&\numprint{113971}&\numprint{122128}&\numprint{143562}&\numprint{120371}&\numprint{100640}\\
8&\textbf{\numprint{11236}}&\numprint{99288}&\numprint{106215}&\numprint{118378}&\numprint{105942}&\numprint{84381}\\
9&\textbf{\numprint{20822}}&\numprint{125358}&\numprint{139004}&\numprint{168837}&\numprint{138395}&\numprint{111245}\\
\midrule
all&\textbf{\numprint{16361}}&\numprint{99490}&\numprint{102410}&\numprint{120936}&\numprint{113587}&\numprint{88805}\\
\bottomrule
\end{tabular}
\end{table}

\begin{table}
\centering
\caption{Mean F-score, precision, and recall (macro-averaging) of different kernels evaluated by means of the Reuters corpus.}
\label{tab:eval_reuters}
\begin{tabular}{llll}
\toprule
Kernel&F-score&Precision&Recall\\
\midrule
NDK&0.394&0.414&0.419\\
Square&0.324&0.392&0.304\\
Cubic&0.348&0.319&\textbf{0.567}\\
RBF&0.403&0.420&0.441\\
Linear&\textbf{0.408}&\textbf{0.428}&0.436\\
\bottomrule
\end{tabular}
\end{table}


\FloatBarrier 

\section*{Acknowledgement}

We thank Vincent Esche and Tom Kollmar for valuable comments and suggestions for improving our paper.
\balance
\bibliographystyle{plain}
\bibliography{powerkernel_icmla2015}

\end{document}

%% file: pics/histogram.tex
\setlength{\unitlength}{0.240900pt}
\ifx\plotpoint\undefined\newsavebox{\plotpoint}\fi
\sbox{\plotpoint}{\rule[-0.200pt]{0.400pt}{0.400pt}}%
\begin{picture}(1050,630)(0,0)
\sbox{\plotpoint}{\rule[-0.200pt]{0.400pt}{0.400pt}}%
\put(150.0,82.0){\rule[-0.200pt]{4.818pt}{0.400pt}}
\put(130,82){\makebox(0,0)[r]{ 0}}
\put(969.0,82.0){\rule[-0.200pt]{4.818pt}{0.400pt}}
\put(150.0,167.0){\rule[-0.200pt]{4.818pt}{0.400pt}}
\put(130,167){\makebox(0,0)[r]{ 500}}
\put(969.0,167.0){\rule[-0.200pt]{4.818pt}{0.400pt}}
\put(150.0,251.0){\rule[-0.200pt]{4.818pt}{0.400pt}}
\put(130,251){\makebox(0,0)[r]{ 1000}}
\put(969.0,251.0){\rule[-0.200pt]{4.818pt}{0.400pt}}
\put(150.0,336.0){\rule[-0.200pt]{4.818pt}{0.400pt}}
\put(130,336){\makebox(0,0)[r]{ 1500}}
\put(969.0,336.0){\rule[-0.200pt]{4.818pt}{0.400pt}}
\put(150.0,420.0){\rule[-0.200pt]{4.818pt}{0.400pt}}
\put(130,420){\makebox(0,0)[r]{ 2000}}
\put(969.0,420.0){\rule[-0.200pt]{4.818pt}{0.400pt}}
\put(150.0,505.0){\rule[-0.200pt]{4.818pt}{0.400pt}}
\put(130,505){\makebox(0,0)[r]{ 2500}}
\put(969.0,505.0){\rule[-0.200pt]{4.818pt}{0.400pt}}
\put(150.0,589.0){\rule[-0.200pt]{4.818pt}{0.400pt}}
\put(130,589){\makebox(0,0)[r]{ 3000}}
\put(969.0,589.0){\rule[-0.200pt]{4.818pt}{0.400pt}}
\put(150.0,82.0){\rule[-0.200pt]{0.400pt}{4.818pt}}
\put(150,41){\makebox(0,0){-10}}
\put(150.0,569.0){\rule[-0.200pt]{0.400pt}{4.818pt}}
\put(226.0,82.0){\rule[-0.200pt]{0.400pt}{4.818pt}}
\put(226,41){\makebox(0,0){ 0}}
\put(226.0,569.0){\rule[-0.200pt]{0.400pt}{4.818pt}}
\put(303.0,82.0){\rule[-0.200pt]{0.400pt}{4.818pt}}
\put(303,41){\makebox(0,0){ 10}}
\put(303.0,569.0){\rule[-0.200pt]{0.400pt}{4.818pt}}
\put(379.0,82.0){\rule[-0.200pt]{0.400pt}{4.818pt}}
\put(379,41){\makebox(0,0){ 20}}
\put(379.0,569.0){\rule[-0.200pt]{0.400pt}{4.818pt}}
\put(455.0,82.0){\rule[-0.200pt]{0.400pt}{4.818pt}}
\put(455,41){\makebox(0,0){ 30}}
\put(455.0,569.0){\rule[-0.200pt]{0.400pt}{4.818pt}}
\put(531.0,82.0){\rule[-0.200pt]{0.400pt}{4.818pt}}
\put(531,41){\makebox(0,0){ 40}}
\put(531.0,569.0){\rule[-0.200pt]{0.400pt}{4.818pt}}
\put(608.0,82.0){\rule[-0.200pt]{0.400pt}{4.818pt}}
\put(608,41){\makebox(0,0){ 50}}
\put(608.0,569.0){\rule[-0.200pt]{0.400pt}{4.818pt}}
\put(684.0,82.0){\rule[-0.200pt]{0.400pt}{4.818pt}}
\put(684,41){\makebox(0,0){ 60}}
\put(684.0,569.0){\rule[-0.200pt]{0.400pt}{4.818pt}}
\put(760.0,82.0){\rule[-0.200pt]{0.400pt}{4.818pt}}
\put(760,41){\makebox(0,0){ 70}}
\put(760.0,569.0){\rule[-0.200pt]{0.400pt}{4.818pt}}
\put(836.0,82.0){\rule[-0.200pt]{0.400pt}{4.818pt}}
\put(836,41){\makebox(0,0){ 80}}
\put(836.0,569.0){\rule[-0.200pt]{0.400pt}{4.818pt}}
\put(913.0,82.0){\rule[-0.200pt]{0.400pt}{4.818pt}}
\put(913,41){\makebox(0,0){ 90}}
\put(913.0,569.0){\rule[-0.200pt]{0.400pt}{4.818pt}}
\put(989.0,82.0){\rule[-0.200pt]{0.400pt}{4.818pt}}
\put(989,41){\makebox(0,0){ 100}}
\put(989.0,569.0){\rule[-0.200pt]{0.400pt}{4.818pt}}
\put(150.0,82.0){\rule[-0.200pt]{0.400pt}{122.136pt}}
\put(150.0,82.0){\rule[-0.200pt]{202.115pt}{0.400pt}}
\put(989.0,82.0){\rule[-0.200pt]{0.400pt}{122.136pt}}
\put(150.0,589.0){\rule[-0.200pt]{202.115pt}{0.400pt}}
\put(829,549){\makebox(0,0)[r]{Number of categories}}
\put(849,539){\rule{24.09pt}{4.818pt}}
\put(849.0,539.0){\rule[-0.200pt]{24.090pt}{0.400pt}}
\put(949.0,539.0){\rule[-0.200pt]{0.400pt}{4.818pt}}
\put(849.0,559.0){\rule[-0.200pt]{24.090pt}{0.400pt}}
\put(849.0,539.0){\rule[-0.200pt]{0.400pt}{4.818pt}}
\put(226,82){\rule{0.9636pt}{117.318pt}}
\put(226.0,82.0){\rule[-0.200pt]{0.400pt}{117.077pt}}
\put(226.0,568.0){\rule[-0.200pt]{0.723pt}{0.400pt}}
\put(229.0,82.0){\rule[-0.200pt]{0.400pt}{117.077pt}}
\put(226.0,82.0){\rule[-0.200pt]{0.723pt}{0.400pt}}
\put(234,82){\rule{0.7227pt}{74.679pt}}
\put(234.0,82.0){\rule[-0.200pt]{0.400pt}{74.438pt}}
\put(234.0,391.0){\rule[-0.200pt]{0.482pt}{0.400pt}}
\put(236.0,82.0){\rule[-0.200pt]{0.400pt}{74.438pt}}
\put(234.0,82.0){\rule[-0.200pt]{0.482pt}{0.400pt}}
\put(242,82){\rule{0.7227pt}{67.452pt}}
\put(242.0,82.0){\rule[-0.200pt]{0.400pt}{67.211pt}}
\put(242.0,361.0){\rule[-0.200pt]{0.482pt}{0.400pt}}
\put(244.0,82.0){\rule[-0.200pt]{0.400pt}{67.211pt}}
\put(242.0,82.0){\rule[-0.200pt]{0.482pt}{0.400pt}}
\put(249,82){\rule{0.9636pt}{22.1628pt}}
\put(249.0,82.0){\rule[-0.200pt]{0.400pt}{21.922pt}}
\put(249.0,173.0){\rule[-0.200pt]{0.723pt}{0.400pt}}
\put(252.0,82.0){\rule[-0.200pt]{0.400pt}{21.922pt}}
\put(249.0,82.0){\rule[-0.200pt]{0.723pt}{0.400pt}}
\put(257,82){\rule{0.7227pt}{17.8266pt}}
\put(257.0,82.0){\rule[-0.200pt]{0.400pt}{17.586pt}}
\put(257.0,155.0){\rule[-0.200pt]{0.482pt}{0.400pt}}
\put(259.0,82.0){\rule[-0.200pt]{0.400pt}{17.586pt}}
\put(257.0,82.0){\rule[-0.200pt]{0.482pt}{0.400pt}}
\put(265,82){\rule{0.7227pt}{16.1403pt}}
\put(265.0,82.0){\rule[-0.200pt]{0.400pt}{15.899pt}}
\put(265.0,148.0){\rule[-0.200pt]{0.482pt}{0.400pt}}
\put(267.0,82.0){\rule[-0.200pt]{0.400pt}{15.899pt}}
\put(265.0,82.0){\rule[-0.200pt]{0.482pt}{0.400pt}}
\put(272,82){\rule{0.7227pt}{15.1767pt}}
\put(272.0,82.0){\rule[-0.200pt]{0.400pt}{14.936pt}}
\put(272.0,144.0){\rule[-0.200pt]{0.482pt}{0.400pt}}
\put(274.0,82.0){\rule[-0.200pt]{0.400pt}{14.936pt}}
\put(272.0,82.0){\rule[-0.200pt]{0.482pt}{0.400pt}}
\put(280,82){\rule{0.7227pt}{14.454pt}}
\put(280.0,82.0){\rule[-0.200pt]{0.400pt}{14.213pt}}
\put(280.0,141.0){\rule[-0.200pt]{0.482pt}{0.400pt}}
\put(282.0,82.0){\rule[-0.200pt]{0.400pt}{14.213pt}}
\put(280.0,82.0){\rule[-0.200pt]{0.482pt}{0.400pt}}
\put(287,82){\rule{0.9636pt}{8.9133pt}}
\put(287.0,82.0){\rule[-0.200pt]{0.400pt}{8.672pt}}
\put(287.0,118.0){\rule[-0.200pt]{0.723pt}{0.400pt}}
\put(290.0,82.0){\rule[-0.200pt]{0.400pt}{8.672pt}}
\put(287.0,82.0){\rule[-0.200pt]{0.723pt}{0.400pt}}
\put(295,82){\rule{0.7227pt}{8.1906pt}}
\put(295.0,82.0){\rule[-0.200pt]{0.400pt}{7.950pt}}
\put(295.0,115.0){\rule[-0.200pt]{0.482pt}{0.400pt}}
\put(297.0,82.0){\rule[-0.200pt]{0.400pt}{7.950pt}}
\put(295.0,82.0){\rule[-0.200pt]{0.482pt}{0.400pt}}
\put(303,82){\rule{0.7227pt}{7.7088pt}}
\put(303.0,82.0){\rule[-0.200pt]{0.400pt}{7.468pt}}
\put(303.0,113.0){\rule[-0.200pt]{0.482pt}{0.400pt}}
\put(305.0,82.0){\rule[-0.200pt]{0.400pt}{7.468pt}}
\put(303.0,82.0){\rule[-0.200pt]{0.482pt}{0.400pt}}
\put(310,82){\rule{0.9636pt}{6.0225pt}}
\put(310.0,82.0){\rule[-0.200pt]{0.400pt}{5.782pt}}
\put(310.0,106.0){\rule[-0.200pt]{0.723pt}{0.400pt}}
\put(313.0,82.0){\rule[-0.200pt]{0.400pt}{5.782pt}}
\put(310.0,82.0){\rule[-0.200pt]{0.723pt}{0.400pt}}
\put(318,82){\rule{0.7227pt}{5.5407pt}}
\put(318.0,82.0){\rule[-0.200pt]{0.400pt}{5.300pt}}
\put(318.0,104.0){\rule[-0.200pt]{0.482pt}{0.400pt}}
\put(320.0,82.0){\rule[-0.200pt]{0.400pt}{5.300pt}}
\put(318.0,82.0){\rule[-0.200pt]{0.482pt}{0.400pt}}
\put(326,82){\rule{0.7227pt}{5.2998pt}}
\put(326.0,82.0){\rule[-0.200pt]{0.400pt}{5.059pt}}
\put(326.0,103.0){\rule[-0.200pt]{0.482pt}{0.400pt}}
\put(328.0,82.0){\rule[-0.200pt]{0.400pt}{5.059pt}}
\put(326.0,82.0){\rule[-0.200pt]{0.482pt}{0.400pt}}
\put(333,82){\rule{0.7227pt}{5.2998pt}}
\put(333.0,82.0){\rule[-0.200pt]{0.400pt}{5.059pt}}
\put(333.0,103.0){\rule[-0.200pt]{0.482pt}{0.400pt}}
\put(335.0,82.0){\rule[-0.200pt]{0.400pt}{5.059pt}}
\put(333.0,82.0){\rule[-0.200pt]{0.482pt}{0.400pt}}
\put(341,82){\rule{0.7227pt}{4.818pt}}
\put(341.0,82.0){\rule[-0.200pt]{0.400pt}{4.577pt}}
\put(341.0,101.0){\rule[-0.200pt]{0.482pt}{0.400pt}}
\put(343.0,82.0){\rule[-0.200pt]{0.400pt}{4.577pt}}
\put(341.0,82.0){\rule[-0.200pt]{0.482pt}{0.400pt}}
\put(348,82){\rule{0.9636pt}{4.3362pt}}
\put(348.0,82.0){\rule[-0.200pt]{0.400pt}{4.095pt}}
\put(348.0,99.0){\rule[-0.200pt]{0.723pt}{0.400pt}}
\put(351.0,82.0){\rule[-0.200pt]{0.400pt}{4.095pt}}
\put(348.0,82.0){\rule[-0.200pt]{0.723pt}{0.400pt}}
\put(356,82){\rule{0.7227pt}{4.0953pt}}
\put(356.0,82.0){\rule[-0.200pt]{0.400pt}{3.854pt}}
\put(356.0,98.0){\rule[-0.200pt]{0.482pt}{0.400pt}}
\put(358.0,82.0){\rule[-0.200pt]{0.400pt}{3.854pt}}
\put(356.0,82.0){\rule[-0.200pt]{0.482pt}{0.400pt}}
\put(364,82){\rule{0.7227pt}{3.8544pt}}
\put(364.0,82.0){\rule[-0.200pt]{0.400pt}{3.613pt}}
\put(364.0,97.0){\rule[-0.200pt]{0.482pt}{0.400pt}}
\put(366.0,82.0){\rule[-0.200pt]{0.400pt}{3.613pt}}
\put(364.0,82.0){\rule[-0.200pt]{0.482pt}{0.400pt}}
\put(371,82){\rule{0.9636pt}{3.3726pt}}
\put(371.0,82.0){\rule[-0.200pt]{0.400pt}{3.132pt}}
\put(371.0,95.0){\rule[-0.200pt]{0.723pt}{0.400pt}}
\put(374.0,82.0){\rule[-0.200pt]{0.400pt}{3.132pt}}
\put(371.0,82.0){\rule[-0.200pt]{0.723pt}{0.400pt}}
\put(379,82){\rule{0.7227pt}{3.3726pt}}
\put(379.0,82.0){\rule[-0.200pt]{0.400pt}{3.132pt}}
\put(379.0,95.0){\rule[-0.200pt]{0.482pt}{0.400pt}}
\put(381.0,82.0){\rule[-0.200pt]{0.400pt}{3.132pt}}
\put(379.0,82.0){\rule[-0.200pt]{0.482pt}{0.400pt}}
\put(387,82){\rule{0.7227pt}{3.3726pt}}
\put(387.0,82.0){\rule[-0.200pt]{0.400pt}{3.132pt}}
\put(387.0,95.0){\rule[-0.200pt]{0.482pt}{0.400pt}}
\put(389.0,82.0){\rule[-0.200pt]{0.400pt}{3.132pt}}
\put(387.0,82.0){\rule[-0.200pt]{0.482pt}{0.400pt}}
\put(394,82){\rule{0.7227pt}{3.3726pt}}
\put(394.0,82.0){\rule[-0.200pt]{0.400pt}{3.132pt}}
\put(394.0,95.0){\rule[-0.200pt]{0.482pt}{0.400pt}}
\put(396.0,82.0){\rule[-0.200pt]{0.400pt}{3.132pt}}
\put(394.0,82.0){\rule[-0.200pt]{0.482pt}{0.400pt}}
\put(402,82){\rule{0.7227pt}{3.1317pt}}
\put(402.0,82.0){\rule[-0.200pt]{0.400pt}{2.891pt}}
\put(402.0,94.0){\rule[-0.200pt]{0.482pt}{0.400pt}}
\put(404.0,82.0){\rule[-0.200pt]{0.400pt}{2.891pt}}
\put(402.0,82.0){\rule[-0.200pt]{0.482pt}{0.400pt}}
\put(409,82){\rule{0.9636pt}{2.409pt}}
\put(409.0,82.0){\rule[-0.200pt]{0.400pt}{2.168pt}}
\put(409.0,91.0){\rule[-0.200pt]{0.723pt}{0.400pt}}
\put(412.0,82.0){\rule[-0.200pt]{0.400pt}{2.168pt}}
\put(409.0,82.0){\rule[-0.200pt]{0.723pt}{0.400pt}}
\put(417,82){\rule{0.7227pt}{2.409pt}}
\put(417.0,82.0){\rule[-0.200pt]{0.400pt}{2.168pt}}
\put(417.0,91.0){\rule[-0.200pt]{0.482pt}{0.400pt}}
\put(419.0,82.0){\rule[-0.200pt]{0.400pt}{2.168pt}}
\put(417.0,82.0){\rule[-0.200pt]{0.482pt}{0.400pt}}
\put(425,82){\rule{0.7227pt}{2.1681pt}}
\put(425.0,82.0){\rule[-0.200pt]{0.400pt}{1.927pt}}
\put(425.0,90.0){\rule[-0.200pt]{0.482pt}{0.400pt}}
\put(427.0,82.0){\rule[-0.200pt]{0.400pt}{1.927pt}}
\put(425.0,82.0){\rule[-0.200pt]{0.482pt}{0.400pt}}
\put(432,82){\rule{0.9636pt}{2.1681pt}}
\put(432.0,82.0){\rule[-0.200pt]{0.400pt}{1.927pt}}
\put(432.0,90.0){\rule[-0.200pt]{0.723pt}{0.400pt}}
\put(435.0,82.0){\rule[-0.200pt]{0.400pt}{1.927pt}}
\put(432.0,82.0){\rule[-0.200pt]{0.723pt}{0.400pt}}
\put(440,82){\rule{0.7227pt}{2.1681pt}}
\put(440.0,82.0){\rule[-0.200pt]{0.400pt}{1.927pt}}
\put(440.0,90.0){\rule[-0.200pt]{0.482pt}{0.400pt}}
\put(442.0,82.0){\rule[-0.200pt]{0.400pt}{1.927pt}}
\put(440.0,82.0){\rule[-0.200pt]{0.482pt}{0.400pt}}
\put(448,82){\rule{0.7227pt}{2.1681pt}}
\put(448.0,82.0){\rule[-0.200pt]{0.400pt}{1.927pt}}
\put(448.0,90.0){\rule[-0.200pt]{0.482pt}{0.400pt}}
\put(450.0,82.0){\rule[-0.200pt]{0.400pt}{1.927pt}}
\put(448.0,82.0){\rule[-0.200pt]{0.482pt}{0.400pt}}
\put(455,82){\rule{0.9636pt}{1.9272pt}}
\put(455.0,82.0){\rule[-0.200pt]{0.400pt}{1.686pt}}
\put(455.0,89.0){\rule[-0.200pt]{0.723pt}{0.400pt}}
\put(458.0,82.0){\rule[-0.200pt]{0.400pt}{1.686pt}}
\put(455.0,82.0){\rule[-0.200pt]{0.723pt}{0.400pt}}
\put(463,82){\rule{0.7227pt}{1.9272pt}}
\put(463.0,82.0){\rule[-0.200pt]{0.400pt}{1.686pt}}
\put(463.0,89.0){\rule[-0.200pt]{0.482pt}{0.400pt}}
\put(465.0,82.0){\rule[-0.200pt]{0.400pt}{1.686pt}}
\put(463.0,82.0){\rule[-0.200pt]{0.482pt}{0.400pt}}
\put(470,82){\rule{0.9636pt}{1.9272pt}}
\put(470.0,82.0){\rule[-0.200pt]{0.400pt}{1.686pt}}
\put(470.0,89.0){\rule[-0.200pt]{0.723pt}{0.400pt}}
\put(473.0,82.0){\rule[-0.200pt]{0.400pt}{1.686pt}}
\put(470.0,82.0){\rule[-0.200pt]{0.723pt}{0.400pt}}
\put(478,82){\rule{0.7227pt}{1.6863pt}}
\put(478.0,82.0){\rule[-0.200pt]{0.400pt}{1.445pt}}
\put(478.0,88.0){\rule[-0.200pt]{0.482pt}{0.400pt}}
\put(480.0,82.0){\rule[-0.200pt]{0.400pt}{1.445pt}}
\put(478.0,82.0){\rule[-0.200pt]{0.482pt}{0.400pt}}
\put(486,82){\rule{0.7227pt}{1.6863pt}}
\put(486.0,82.0){\rule[-0.200pt]{0.400pt}{1.445pt}}
\put(486.0,88.0){\rule[-0.200pt]{0.482pt}{0.400pt}}
\put(488.0,82.0){\rule[-0.200pt]{0.400pt}{1.445pt}}
\put(486.0,82.0){\rule[-0.200pt]{0.482pt}{0.400pt}}
\put(493,82){\rule{0.9636pt}{1.6863pt}}
\put(493.0,82.0){\rule[-0.200pt]{0.400pt}{1.445pt}}
\put(493.0,88.0){\rule[-0.200pt]{0.723pt}{0.400pt}}
\put(496.0,82.0){\rule[-0.200pt]{0.400pt}{1.445pt}}
\put(493.0,82.0){\rule[-0.200pt]{0.723pt}{0.400pt}}
\put(501,82){\rule{0.7227pt}{1.6863pt}}
\put(501.0,82.0){\rule[-0.200pt]{0.400pt}{1.445pt}}
\put(501.0,88.0){\rule[-0.200pt]{0.482pt}{0.400pt}}
\put(503.0,82.0){\rule[-0.200pt]{0.400pt}{1.445pt}}
\put(501.0,82.0){\rule[-0.200pt]{0.482pt}{0.400pt}}
\put(509,82){\rule{0.7227pt}{1.6863pt}}
\put(509.0,82.0){\rule[-0.200pt]{0.400pt}{1.445pt}}
\put(509.0,88.0){\rule[-0.200pt]{0.482pt}{0.400pt}}
\put(511.0,82.0){\rule[-0.200pt]{0.400pt}{1.445pt}}
\put(509.0,82.0){\rule[-0.200pt]{0.482pt}{0.400pt}}
\put(516,82){\rule{0.9636pt}{1.4454pt}}
\put(516.0,82.0){\rule[-0.200pt]{0.400pt}{1.204pt}}
\put(516.0,87.0){\rule[-0.200pt]{0.723pt}{0.400pt}}
\put(519.0,82.0){\rule[-0.200pt]{0.400pt}{1.204pt}}
\put(516.0,82.0){\rule[-0.200pt]{0.723pt}{0.400pt}}
\put(524,82){\rule{0.7227pt}{1.4454pt}}
\put(524.0,82.0){\rule[-0.200pt]{0.400pt}{1.204pt}}
\put(524.0,87.0){\rule[-0.200pt]{0.482pt}{0.400pt}}
\put(526.0,82.0){\rule[-0.200pt]{0.400pt}{1.204pt}}
\put(524.0,82.0){\rule[-0.200pt]{0.482pt}{0.400pt}}
\put(531,82){\rule{0.9636pt}{1.2045pt}}
\put(531.0,82.0){\rule[-0.200pt]{0.400pt}{0.964pt}}
\put(531.0,86.0){\rule[-0.200pt]{0.723pt}{0.400pt}}
\put(534.0,82.0){\rule[-0.200pt]{0.400pt}{0.964pt}}
\put(531.0,82.0){\rule[-0.200pt]{0.723pt}{0.400pt}}
\put(539,82){\rule{0.7227pt}{1.2045pt}}
\put(539.0,82.0){\rule[-0.200pt]{0.400pt}{0.964pt}}
\put(539.0,86.0){\rule[-0.200pt]{0.482pt}{0.400pt}}
\put(541.0,82.0){\rule[-0.200pt]{0.400pt}{0.964pt}}
\put(539.0,82.0){\rule[-0.200pt]{0.482pt}{0.400pt}}
\put(547,82){\rule{0.7227pt}{1.2045pt}}
\put(547.0,82.0){\rule[-0.200pt]{0.400pt}{0.964pt}}
\put(547.0,86.0){\rule[-0.200pt]{0.482pt}{0.400pt}}
\put(549.0,82.0){\rule[-0.200pt]{0.400pt}{0.964pt}}
\put(547.0,82.0){\rule[-0.200pt]{0.482pt}{0.400pt}}
\put(554,82){\rule{0.9636pt}{1.2045pt}}
\put(554.0,82.0){\rule[-0.200pt]{0.400pt}{0.964pt}}
\put(554.0,86.0){\rule[-0.200pt]{0.723pt}{0.400pt}}
\put(557.0,82.0){\rule[-0.200pt]{0.400pt}{0.964pt}}
\put(554.0,82.0){\rule[-0.200pt]{0.723pt}{0.400pt}}
\put(562,82){\rule{0.7227pt}{0.9636pt}}
\put(562.0,82.0){\rule[-0.200pt]{0.400pt}{0.723pt}}
\put(562.0,85.0){\rule[-0.200pt]{0.482pt}{0.400pt}}
\put(564.0,82.0){\rule[-0.200pt]{0.400pt}{0.723pt}}
\put(562.0,82.0){\rule[-0.200pt]{0.482pt}{0.400pt}}
\put(570,82){\rule{0.7227pt}{0.9636pt}}
\put(570.0,82.0){\rule[-0.200pt]{0.400pt}{0.723pt}}
\put(570.0,85.0){\rule[-0.200pt]{0.482pt}{0.400pt}}
\put(572.0,82.0){\rule[-0.200pt]{0.400pt}{0.723pt}}
\put(570.0,82.0){\rule[-0.200pt]{0.482pt}{0.400pt}}
\put(577,82){\rule{0.9636pt}{0.9636pt}}
\put(577.0,82.0){\rule[-0.200pt]{0.400pt}{0.723pt}}
\put(577.0,85.0){\rule[-0.200pt]{0.723pt}{0.400pt}}
\put(580.0,82.0){\rule[-0.200pt]{0.400pt}{0.723pt}}
\put(577.0,82.0){\rule[-0.200pt]{0.723pt}{0.400pt}}
\put(585,82){\rule{0.7227pt}{0.9636pt}}
\put(585.0,82.0){\rule[-0.200pt]{0.400pt}{0.723pt}}
\put(585.0,85.0){\rule[-0.200pt]{0.482pt}{0.400pt}}
\put(587.0,82.0){\rule[-0.200pt]{0.400pt}{0.723pt}}
\put(585.0,82.0){\rule[-0.200pt]{0.482pt}{0.400pt}}
\put(593,82){\rule{0.7227pt}{0.9636pt}}
\put(593.0,82.0){\rule[-0.200pt]{0.400pt}{0.723pt}}
\put(593.0,85.0){\rule[-0.200pt]{0.482pt}{0.400pt}}
\put(595.0,82.0){\rule[-0.200pt]{0.400pt}{0.723pt}}
\put(593.0,82.0){\rule[-0.200pt]{0.482pt}{0.400pt}}
\put(600,82){\rule{0.7227pt}{0.9636pt}}
\put(600.0,82.0){\rule[-0.200pt]{0.400pt}{0.723pt}}
\put(600.0,85.0){\rule[-0.200pt]{0.482pt}{0.400pt}}
\put(602.0,82.0){\rule[-0.200pt]{0.400pt}{0.723pt}}
\put(600.0,82.0){\rule[-0.200pt]{0.482pt}{0.400pt}}
\put(608,82){\rule{0.7227pt}{0.9636pt}}
\put(608.0,82.0){\rule[-0.200pt]{0.400pt}{0.723pt}}
\put(608.0,85.0){\rule[-0.200pt]{0.482pt}{0.400pt}}
\put(610.0,82.0){\rule[-0.200pt]{0.400pt}{0.723pt}}
\put(608.0,82.0){\rule[-0.200pt]{0.482pt}{0.400pt}}
\put(615,82){\rule{0.9636pt}{0.9636pt}}
\put(615.0,82.0){\rule[-0.200pt]{0.400pt}{0.723pt}}
\put(615.0,85.0){\rule[-0.200pt]{0.723pt}{0.400pt}}
\put(618.0,82.0){\rule[-0.200pt]{0.400pt}{0.723pt}}
\put(615.0,82.0){\rule[-0.200pt]{0.723pt}{0.400pt}}
\put(623,82){\rule{0.7227pt}{0.9636pt}}
\put(623.0,82.0){\rule[-0.200pt]{0.400pt}{0.723pt}}
\put(623.0,85.0){\rule[-0.200pt]{0.482pt}{0.400pt}}
\put(625.0,82.0){\rule[-0.200pt]{0.400pt}{0.723pt}}
\put(623.0,82.0){\rule[-0.200pt]{0.482pt}{0.400pt}}
\put(631,82){\rule{0.7227pt}{0.7227pt}}
\put(631.0,82.0){\rule[-0.200pt]{0.400pt}{0.482pt}}
\put(631.0,84.0){\rule[-0.200pt]{0.482pt}{0.400pt}}
\put(633.0,82.0){\rule[-0.200pt]{0.400pt}{0.482pt}}
\put(631.0,82.0){\rule[-0.200pt]{0.482pt}{0.400pt}}
\put(638,82){\rule{0.9636pt}{0.7227pt}}
\put(638.0,82.0){\rule[-0.200pt]{0.400pt}{0.482pt}}
\put(638.0,84.0){\rule[-0.200pt]{0.723pt}{0.400pt}}
\put(641.0,82.0){\rule[-0.200pt]{0.400pt}{0.482pt}}
\put(638.0,82.0){\rule[-0.200pt]{0.723pt}{0.400pt}}
\put(646,82){\rule{0.7227pt}{0.7227pt}}
\put(646.0,82.0){\rule[-0.200pt]{0.400pt}{0.482pt}}
\put(646.0,84.0){\rule[-0.200pt]{0.482pt}{0.400pt}}
\put(648.0,82.0){\rule[-0.200pt]{0.400pt}{0.482pt}}
\put(646.0,82.0){\rule[-0.200pt]{0.482pt}{0.400pt}}
\put(654,82){\rule{0.7227pt}{0.7227pt}}
\put(654.0,82.0){\rule[-0.200pt]{0.400pt}{0.482pt}}
\put(654.0,84.0){\rule[-0.200pt]{0.482pt}{0.400pt}}
\put(656.0,82.0){\rule[-0.200pt]{0.400pt}{0.482pt}}
\put(654.0,82.0){\rule[-0.200pt]{0.482pt}{0.400pt}}
\put(661,82){\rule{0.7227pt}{0.7227pt}}
\put(661.0,82.0){\rule[-0.200pt]{0.400pt}{0.482pt}}
\put(661.0,84.0){\rule[-0.200pt]{0.482pt}{0.400pt}}
\put(663.0,82.0){\rule[-0.200pt]{0.400pt}{0.482pt}}
\put(661.0,82.0){\rule[-0.200pt]{0.482pt}{0.400pt}}
\put(669,82){\rule{0.7227pt}{0.7227pt}}
\put(669.0,82.0){\rule[-0.200pt]{0.400pt}{0.482pt}}
\put(669.0,84.0){\rule[-0.200pt]{0.482pt}{0.400pt}}
\put(671.0,82.0){\rule[-0.200pt]{0.400pt}{0.482pt}}
\put(669.0,82.0){\rule[-0.200pt]{0.482pt}{0.400pt}}
\put(676,82){\rule{0.9636pt}{0.7227pt}}
\put(676.0,82.0){\rule[-0.200pt]{0.400pt}{0.482pt}}
\put(676.0,84.0){\rule[-0.200pt]{0.723pt}{0.400pt}}
\put(679.0,82.0){\rule[-0.200pt]{0.400pt}{0.482pt}}
\put(676.0,82.0){\rule[-0.200pt]{0.723pt}{0.400pt}}
\put(684,82){\rule{0.7227pt}{0.7227pt}}
\put(684.0,82.0){\rule[-0.200pt]{0.400pt}{0.482pt}}
\put(684.0,84.0){\rule[-0.200pt]{0.482pt}{0.400pt}}
\put(686.0,82.0){\rule[-0.200pt]{0.400pt}{0.482pt}}
\put(684.0,82.0){\rule[-0.200pt]{0.482pt}{0.400pt}}
\put(692,82){\rule{0.7227pt}{0.7227pt}}
\put(692.0,82.0){\rule[-0.200pt]{0.400pt}{0.482pt}}
\put(692.0,84.0){\rule[-0.200pt]{0.482pt}{0.400pt}}
\put(694.0,82.0){\rule[-0.200pt]{0.400pt}{0.482pt}}
\put(692.0,82.0){\rule[-0.200pt]{0.482pt}{0.400pt}}
\put(699,82){\rule{0.9636pt}{0.7227pt}}
\put(699.0,82.0){\rule[-0.200pt]{0.400pt}{0.482pt}}
\put(699.0,84.0){\rule[-0.200pt]{0.723pt}{0.400pt}}
\put(702.0,82.0){\rule[-0.200pt]{0.400pt}{0.482pt}}
\put(699.0,82.0){\rule[-0.200pt]{0.723pt}{0.400pt}}
\put(707,82){\rule{0.7227pt}{0.4818pt}}
\put(707.0,82.0){\usebox{\plotpoint}}
\put(707.0,83.0){\rule[-0.200pt]{0.482pt}{0.400pt}}
\put(709.0,82.0){\usebox{\plotpoint}}
\put(707.0,82.0){\rule[-0.200pt]{0.482pt}{0.400pt}}
\put(715,82){\rule{0.7227pt}{0.4818pt}}
\put(715.0,82.0){\usebox{\plotpoint}}
\put(715.0,83.0){\rule[-0.200pt]{0.482pt}{0.400pt}}
\put(717.0,82.0){\usebox{\plotpoint}}
\put(715.0,82.0){\rule[-0.200pt]{0.482pt}{0.400pt}}
\put(722,82){\rule{0.7227pt}{0.4818pt}}
\put(722.0,82.0){\usebox{\plotpoint}}
\put(722.0,83.0){\rule[-0.200pt]{0.482pt}{0.400pt}}
\put(724.0,82.0){\usebox{\plotpoint}}
\put(722.0,82.0){\rule[-0.200pt]{0.482pt}{0.400pt}}
\put(730,82){\rule{0.7227pt}{0.4818pt}}
\put(730.0,82.0){\usebox{\plotpoint}}
\put(730.0,83.0){\rule[-0.200pt]{0.482pt}{0.400pt}}
\put(732.0,82.0){\usebox{\plotpoint}}
\put(730.0,82.0){\rule[-0.200pt]{0.482pt}{0.400pt}}
\put(737,82){\rule{0.9636pt}{0.4818pt}}
\put(737.0,82.0){\usebox{\plotpoint}}
\put(737.0,83.0){\rule[-0.200pt]{0.723pt}{0.400pt}}
\put(740.0,82.0){\usebox{\plotpoint}}
\put(737.0,82.0){\rule[-0.200pt]{0.723pt}{0.400pt}}
\put(745,82){\rule{0.7227pt}{0.4818pt}}
\put(745.0,82.0){\usebox{\plotpoint}}
\put(745.0,83.0){\rule[-0.200pt]{0.482pt}{0.400pt}}
\put(747.0,82.0){\usebox{\plotpoint}}
\put(745.0,82.0){\rule[-0.200pt]{0.482pt}{0.400pt}}
\put(753,82){\rule{0.7227pt}{0.4818pt}}
\put(753.0,82.0){\usebox{\plotpoint}}
\put(753.0,83.0){\rule[-0.200pt]{0.482pt}{0.400pt}}
\put(755.0,82.0){\usebox{\plotpoint}}
\put(753.0,82.0){\rule[-0.200pt]{0.482pt}{0.400pt}}
\put(760,82){\rule{0.9636pt}{0.4818pt}}
\put(760.0,82.0){\usebox{\plotpoint}}
\put(760.0,83.0){\rule[-0.200pt]{0.723pt}{0.400pt}}
\put(763.0,82.0){\usebox{\plotpoint}}
\put(760.0,82.0){\rule[-0.200pt]{0.723pt}{0.400pt}}
\put(768,82){\rule{0.7227pt}{0.4818pt}}
\put(768.0,82.0){\usebox{\plotpoint}}
\put(768.0,83.0){\rule[-0.200pt]{0.482pt}{0.400pt}}
\put(770.0,82.0){\usebox{\plotpoint}}
\put(768.0,82.0){\rule[-0.200pt]{0.482pt}{0.400pt}}
\put(776,82){\rule{0.7227pt}{0.4818pt}}
\put(776.0,82.0){\usebox{\plotpoint}}
\put(776.0,83.0){\rule[-0.200pt]{0.482pt}{0.400pt}}
\put(778.0,82.0){\usebox{\plotpoint}}
\put(776.0,82.0){\rule[-0.200pt]{0.482pt}{0.400pt}}
\put(783,82){\rule{0.7227pt}{0.4818pt}}
\put(783.0,82.0){\usebox{\plotpoint}}
\put(783.0,83.0){\rule[-0.200pt]{0.482pt}{0.400pt}}
\put(785.0,82.0){\usebox{\plotpoint}}
\put(783.0,82.0){\rule[-0.200pt]{0.482pt}{0.400pt}}
\put(791,82){\rule{0.7227pt}{0.4818pt}}
\put(791.0,82.0){\usebox{\plotpoint}}
\put(791.0,83.0){\rule[-0.200pt]{0.482pt}{0.400pt}}
\put(793.0,82.0){\usebox{\plotpoint}}
\put(791.0,82.0){\rule[-0.200pt]{0.482pt}{0.400pt}}
\put(798,82){\rule{0.9636pt}{0.4818pt}}
\put(798.0,82.0){\usebox{\plotpoint}}
\put(798.0,83.0){\rule[-0.200pt]{0.723pt}{0.400pt}}
\put(801.0,82.0){\usebox{\plotpoint}}
\put(798.0,82.0){\rule[-0.200pt]{0.723pt}{0.400pt}}
\put(806,82){\rule{0.7227pt}{0.4818pt}}
\put(806.0,82.0){\usebox{\plotpoint}}
\put(806.0,83.0){\rule[-0.200pt]{0.482pt}{0.400pt}}
\put(808.0,82.0){\usebox{\plotpoint}}
\put(806.0,82.0){\rule[-0.200pt]{0.482pt}{0.400pt}}
\put(814,82){\rule{0.7227pt}{0.2409pt}}
\put(814,82){\usebox{\plotpoint}}
\put(814.0,82.0){\rule[-0.200pt]{0.482pt}{0.400pt}}
\put(814.0,82.0){\rule[-0.200pt]{0.482pt}{0.400pt}}
\put(821,82){\rule{0.9636pt}{0.2409pt}}
\put(821,82){\usebox{\plotpoint}}
\put(821.0,82.0){\rule[-0.200pt]{0.723pt}{0.400pt}}
\put(821.0,82.0){\rule[-0.200pt]{0.723pt}{0.400pt}}
\put(829,82){\rule{0.7227pt}{0.2409pt}}
\put(829,82){\usebox{\plotpoint}}
\put(829.0,82.0){\rule[-0.200pt]{0.482pt}{0.400pt}}
\put(829.0,82.0){\rule[-0.200pt]{0.482pt}{0.400pt}}
\put(837,82){\rule{0.7227pt}{0.2409pt}}
\put(837,82){\usebox{\plotpoint}}
\put(837.0,82.0){\rule[-0.200pt]{0.482pt}{0.400pt}}
\put(837.0,82.0){\rule[-0.200pt]{0.482pt}{0.400pt}}
\put(844,82){\rule{0.7227pt}{0.2409pt}}
\put(844,82){\usebox{\plotpoint}}
\put(844.0,82.0){\rule[-0.200pt]{0.482pt}{0.400pt}}
\put(844.0,82.0){\rule[-0.200pt]{0.482pt}{0.400pt}}
\put(852,82){\rule{0.7227pt}{0.2409pt}}
\put(852,82){\usebox{\plotpoint}}
\put(852.0,82.0){\rule[-0.200pt]{0.482pt}{0.400pt}}
\put(852.0,82.0){\rule[-0.200pt]{0.482pt}{0.400pt}}
\put(859,82){\rule{0.9636pt}{0.2409pt}}
\put(859,82){\usebox{\plotpoint}}
\put(859.0,82.0){\rule[-0.200pt]{0.723pt}{0.400pt}}
\put(859.0,82.0){\rule[-0.200pt]{0.723pt}{0.400pt}}
\put(867,82){\rule{0.7227pt}{0.2409pt}}
\put(867,82){\usebox{\plotpoint}}
\put(867.0,82.0){\rule[-0.200pt]{0.482pt}{0.400pt}}
\put(867.0,82.0){\rule[-0.200pt]{0.482pt}{0.400pt}}
\put(875,82){\rule{0.7227pt}{0.2409pt}}
\put(875,82){\usebox{\plotpoint}}
\put(875.0,82.0){\rule[-0.200pt]{0.482pt}{0.400pt}}
\put(875.0,82.0){\rule[-0.200pt]{0.482pt}{0.400pt}}
\put(882,82){\rule{0.9636pt}{0.2409pt}}
\put(882,82){\usebox{\plotpoint}}
\put(882.0,82.0){\rule[-0.200pt]{0.723pt}{0.400pt}}
\put(882.0,82.0){\rule[-0.200pt]{0.723pt}{0.400pt}}
\put(890,82){\rule{0.7227pt}{0.2409pt}}
\put(890,82){\usebox{\plotpoint}}
\put(890.0,82.0){\rule[-0.200pt]{0.482pt}{0.400pt}}
\put(890.0,82.0){\rule[-0.200pt]{0.482pt}{0.400pt}}
\put(898,82){\rule{0.7227pt}{0.2409pt}}
\put(898,82){\usebox{\plotpoint}}
\put(898.0,82.0){\rule[-0.200pt]{0.482pt}{0.400pt}}
\put(898.0,82.0){\rule[-0.200pt]{0.482pt}{0.400pt}}
\put(905,82){\rule{0.9636pt}{0.2409pt}}
\put(905,82){\usebox{\plotpoint}}
\put(905.0,82.0){\rule[-0.200pt]{0.723pt}{0.400pt}}
\put(905.0,82.0){\rule[-0.200pt]{0.723pt}{0.400pt}}
\put(913,82){\rule{0.7227pt}{0.2409pt}}
\put(913,82){\usebox{\plotpoint}}
\put(913.0,82.0){\rule[-0.200pt]{0.482pt}{0.400pt}}
\put(913.0,82.0){\rule[-0.200pt]{0.482pt}{0.400pt}}
\put(150.0,82.0){\rule[-0.200pt]{0.400pt}{122.136pt}}
\put(150.0,82.0){\rule[-0.200pt]{202.115pt}{0.400pt}}
\put(989.0,82.0){\rule[-0.200pt]{0.400pt}{122.136pt}}
\put(150.0,589.0){\rule[-0.200pt]{202.115pt}{0.400pt}}
\end{picture}